%% file: acl2020.tex
\documentclass[11pt,a4paper]{article}
\pdfoutput=1
\usepackage[draft]{hyperref}
\usepackage{acl2020}
\usepackage{times}
\usepackage{latexsym}
\usepackage{amsmath}
\usepackage{color}
\usepackage{soul}
\usepackage{longtable}
\usepackage{graphicx}
\usepackage{tabularx}
\usepackage{soul}
\usepackage{color}
\usepackage{lipsum,graphicx,subcaption}

\usepackage[latin1]{inputenc}
\usepackage[nodisplayskipstretch]{setspace}

\usepackage{microtype}
\usepackage{multirow,booktabs,blindtext}
\aclfinalcopy 

\newcommand{\nb}[1]{\textcolor{blue}{[NB: #1]}}

\newcommand{\eat}[1]{}

\title{Modeling Label Semantics for Predicting Emotional Reactions}

\author{
  Radhika Gaonkar, Heeyoung Kwon, Mohaddeseh Bastan, Niranjan Balasubramanian \\
  Stony Brook University, Stony Brook, New York \\
  \texttt{\{rgaonkar, heekwon, mbastan, niranjan\}@cs.stonybrook.edu} \\\AND
  Nathanael Chambers \\
  US Naval Academy, Annapolis, MD \\
  \texttt{nchamber@usna.edu}}

\date{}

\begin{document}
\maketitle
\begin{abstract}

Predicting how events induce emotions in the characters of a story is typically seen as a standard multi-label classification task, which usually treats labels as anonymous classes to predict. They ignore information that may be conveyed by the emotion labels themselves. 
We propose that the semantics of emotion labels can guide a model's attention when representing the input story. Further, we observe that the emotions evoked by an event are often related: an event that evokes joy is unlikely to also evoke sadness.
In this work, we explicitly model label classes via label embeddings, and add mechanisms that track label-label correlations both during training and inference. We also introduce a new semi-supervision strategy that regularizes for the correlations on unlabeled data. Our empirical evaluations show that modeling label semantics yields consistent benefits, and we advance the state-of-the-art on an emotion inference task.

\end{abstract}

\input{acl2020-templates/introduction}
\input{acl2020-templates/label_semantic}
\input{acl2020-templates/experimental-setup}
\input{acl2020-templates/results}

\input{acl2020-templates/related}
\input{acl2020-templates/conclusions}
\section*{Acknowledgments}
This work was supported in part by the National Science Foundation under Grant IIS-1617969. This material is also based on research that is in part supported by the Air Force Research Laboratory (AFRL), DARPA, for the KAIROS program under agreement number FA8750-19-2-1003. The U.S. Government is authorized to reproduce and distribute reprints for Governmental purposes notwithstanding any copyright notation thereon. 

\bibliographystyle{acl_natbib}
\bibliography{acl2020}

\end{document}

%% file: acl2020-templates/introduction.tex
\section{Introduction}
Understanding how events in a story affect the characters involved is an integral part of narrative understanding. \citet{rashkin2018modeling} introduced an emotion inference task on a subset of the ROCStories dataset \cite{mostafazadeh2016corpus}, labeling entities with the emotions they experience from the short story contexts. Previous work on this and related tasks typically frame them as multi-label classification problems. The standard approach uses an encoder that produces a representation of the target event along with the surrounding story events, and then pushes it through a classification layer to predict the possible emotion labels~\cite{rashkin2018modeling,wang2018joint}.\\
This classification framework ignores the semantics of the emotions themselves. Each emotion label (e.g., joy) is just a binary prediction. However, consider the sentence, ``Danielle was really short on money". The emotional reaction is \texttt{FEAR} of being short on money. First, if a model had lexical foreknowledge of ``fear", we should expect an improved ability to decide if a target event evokes \texttt{FEAR}. Second, such a model might represent relationships between the emotions themselves. For example, an event that evokes \texttt{FEAR} is likely to evoke \texttt{SADNESS} and unlikely to evoke \texttt{JOY}. When previous models frame this as binary label prediction, they miss out on ways to leverage label semantics.

In this work, we show that explicitly modeling label semantics improves emotion inference. We describe three main contributions\footnote{https://github.com/StonyBrookNLP/emotion-label-semantics}. First, we show how to use embeddings as the label semantics representation. We then propose a label attention network that produces label-informed representations of the event and the story context to improve prediction accuracy. Second, we add mechanisms that can make use of label-label correlations as part of both training and inference. During training, the correlations are used to add a regularization loss. During inference, the prediction logits for each label are modified to incorporate the correlations, thus allowing the model's confidence on one label to influence its prediction of other labels. Third, we show that the label correlations can be used as a semi-supervised signal on the unlabeled portion of the ROCStories dataset.\\
Our empirical evaluations show that adding label semantics consistently improves prediction accuracy, and produces labelings that are more consistent than models without label semantics. Our best model outperforms previously reported results and achieves more than 4.9 points absolute improvement over the BERT classification model yielding a new state-of-the-art result for this task.

%% file: acl2020-templates/label_semantic.tex
\section{Emotion Inference}
The emotion inference task introduced by \citet{rashkin2018modeling} is defined over a subset of short stories from the ROCStories dataset \cite{mostafazadeh2016corpus}. It infers the reactions that each event evokes in the characters of the story, given the story context thus far. For each sentence (i.e. event) in a story, the training data includes annotations of eight emotions. Given a sentence $x_s$ denoting a single event in a story, the task is to label the possible emotional reactions that an event evokes in each character in the story. Since an event can evoke multiple reactions, the task is formulated as a multi-label classification problem.

The standard approach to this task has been as follows. For a given character $c$ and the target sentence $x_s$, collect all previous sentences $x_c$ in the story in which the character $c$ is mentioned as the character context. Encode the target sentence, and the character context to obtain a single representation, and use it as input to a multi-label classification layer for prediction. \citet{rashkin2018modeling} benchmark the performance of multiple encoders (see Section~ \ref{sec:baseline}).

We extend this previous work to integrate label semantics into the model by adding label embeddings (Section~\ref{sec:labelembeddings}) and explicitly representing label-label correlations (Section~\ref{sec:labelcorrelations}).

\section{Label Semantics using Embeddings}
\label{sec:labelembeddings}

A simple strategy to model label semantics is to explicitly represent each with an embedding that captures the surface semantics of its label name. Since the emotion labels correspond to actual words (e.g., joy, fear, etc.), we can initialize them with their corresponding word embeddings (learned from a large corpus). We then use these label embeddings in two ways as detailed below.

\subsection{Label Attention Network}
The label embeddings can be used to guide an encoder network to extract emotion-related information from the sentences. We adopted the Label-Embedding Attentive Network (LEAM) architecture to produce label-focused representations~\cite{wang2018joint}. The main idea behind the LEAM model is to compute attention scores between the label and the representations of the tokens in the input that is to be classified\footnote{The original model used LEAM directly on top of Glove embeddings~\cite{wang2018joint}.}. This can then be used to appropriately weight the contributions of each token to the final representations. In this work, we use LEAM to compute an attention matrix computed over the hidden states produced by the encoder and the label embeddings. The encoder used is the BERT features for each token $B_{t}$ in the text and each of the label sentences $J$. The attention matrix is then used to produce a weighted combination of the contextual representations of the input, using the compatibility matrix $H$, as computed in \cite{wang2018joint}. This gives emotion focused representations $y$ to use for classification: 
\begin{equation}
    H = (J^{T}B_{t})\oslash \hat{H}
\end{equation}

\begin{figure}
    
    \includegraphics[width=\linewidth]{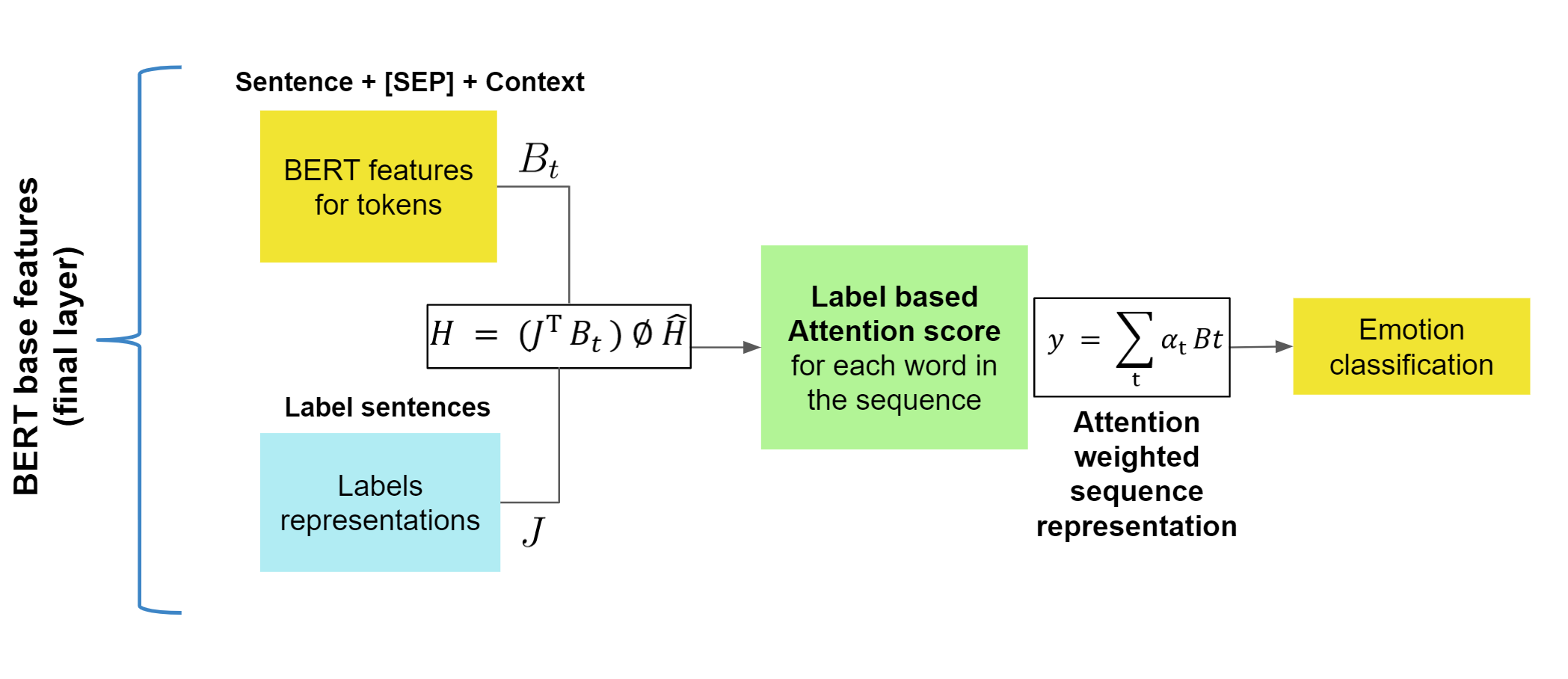}
    \caption{Label-Embedding Attentive Network using BERT Features. $y$ denotes the label attended story sentence and context representation, where $\alpha$ is the attention score.
    }\label{fig:leam_bert}
  \end{figure}

Figure~\ref{fig:leam_bert} illustrates the key steps in the model.
  
\subsection{Labels as Additional Input}
Rather than learning label embeddings from scratch, we also explore using contextual embeddings from transformer-based models like BERT. This allows us to use richer semantics derived from pre-training and also allows us to exploit the self-attention mechanism to introduce label semantics as part of the input itself. In addition to the target and context sentences, we also include emotion-label sentences, $L_s$, of the form ``\texttt{[character] is [emotional state]}'' as input to the classifier. For each instance, we add eight such sentences covering all emotional labels\footnote{This is similar to how answer options are encoded in multiple choice question answering in transformer-based models.}. In this paper, we use the final layer of a pretrained Bert-base model to get representations for the input sentence and each of the emotion-label sentences. The self-attention mechanism will automatically learn to attend to these label sentences when constructing the representations for the input text. 

\section{Label Semantics using Correlations}
\label{sec:labelcorrelations}
\begin{figure}[t]
    
    \includegraphics[width=\linewidth]{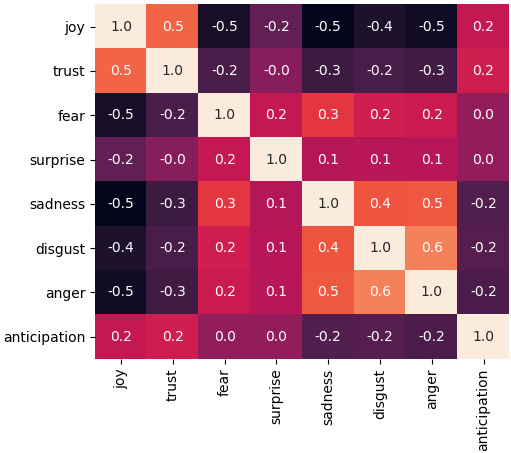}
    \caption{Emotion correlations as seen in the ground truth labels in the training data.
    }\label{fig:label_corrrelations}
  \end{figure}

When more than one emotion is evoked by an event, they aren't independent. Indeed, as shown in Figure~\ref{fig:label_corrrelations}, there are strong (positive and negative) correlations between the emotion labels in the ground truth. For instance, there is a negative correlation ($\rho = -0.5$) between \texttt{JOY} and \texttt{SAD} labels and a positive correlation between \texttt{JOY} and \texttt{TRUST} ($\rho = 0.5$). We propose two ways to incorporate these label correlations to improve prediction.

\subsection{Correlations on Labeled Data}
\label{sec:softtraining}
In a multi-label setting, a good model should respect the label correlations. If it is confident about a particular label, then it should also be confident about other positively correlated labels, and conversely less confident about labels that are negatively correlated. 

Following~\citet{zhao-etal-2019-review}, we add (i) a loss function that penalizes the model for making incongruous predictions, i.e. those that are not compatible with the label correlations, and (ii) a component that multiplies the classification logit vector $z$ with the learned label relations encoded as a learned correlation matrix $G$. This component transforms the raw prediction score of each label to a weighted sum of the prediction scores of the other labels. For each label, these weights are given by its learned correlation with all the other labels. Therefore, the prediction score of each label is affected by the prediction score of the other labels, based on the correlation between label pairs. The final prediction scores are then calculated as shown in the equation:   
\begin{equation}
    e = \sigma(z \cdot G)
    \label{eqn:soft_pred}
\end{equation}

The overall loss then comprises of two loss functions - the prediction loss ($\mathcal{L}_{BCE}$), and the correlation-loss ($\mathcal{L}_{corr}$): 
\begin{equation}
    \mathcal{L}(\theta) = \mathcal{L}_{BCE}(e, y) + \mathcal{L}_{corr}(e, y')
    \label{eqn:loss_soft}
\end{equation}

Where $\mathcal{L}_{corr}$ computes BCE Loss with continuous representation of the true labels $y$, using the learned label correlation $G$:
\begin{equation}
    y' = y \cdot G
\end{equation}

\subsection{Semi-supervision on Unlabeled Data}
We also introduce a new semi-supervision idea to exploit label correlations as a regularization signal on unlabeled data. The multi-label annotations used in this work~\cite{rashkin2018modeling} only comprises a small fraction of the original ROCStories data. There are $\sim$40k character-line pairs that have open text descriptions of emotional reactions, but these aren't annotated with multi-label emotions, and therefore were not used in the above supervised emotion prediction tasks. 
We propose a new semi-supervised method over BERT representations that augments the soft-training objective used in Section \ref{sec:softtraining} with a label correlation incompatibility loss defined over the unlabeled portion of the ROCStories dataset. 

We use two loss functions: the loss computed in Equation \ref{eqn:loss_soft}, and the regularization loss on the unlabeled training data (Equation~\ref{eqn:semi-supervised}).
For the semi-supervised training, we use an iterative batch-wise training. In the first step, all weights of the model are minimized by minimizing the loss in Equation \ref{eqn:loss_soft}. In the next step, the learned label correlations are updated using:

\begin{equation}
      \mathcal{L}_{reg} = \sum_{i, j} G_{ij} \cdot d(e_i, e_j)\\
      \label{eqn:semi-supervised}
\end{equation}

\begin{equation*}
d(e_i, e_j)=
      \begin{cases}
      \rVert e_i - e_j \lVert & \text{for } G_{ij} \geq 0,\\
      \rVert e_i - e_j \lVert - 1 & \text{otherwise.}
      \end{cases}
      \label{eqn:distance}
\end{equation*}
This loss helps the model to produce consistent predictions based on the correlations by forcing positively correlated labels to have similar scores and negatively correlated ones to have dissimilar scores.

%% file: acl2020-templates/experimental-setup.tex
\begin{table*}[ht!]
    \centering
    
    \begin{tabular}{|c|l|c|c|c|}
    \hline
      & \textbf{Model} & \textbf{Precision} & \textbf{Recall} & \textbf{F1}\\
      \cline{2-5}
      & \citet{rashkin2018modeling} & & & \\ 
      & {BiLSTM} & {25.31} & {33.44} & {28.81}\\ 
      & {CNN} & 24.47 & 38.87 & 30.04\\ 
      Baselines & {REN} & 25.30 & 37.30 & 30.15\\ 
      
      & {NPN} & 24.33 & 40.10 & 30.29\\ 
      \cline{2-5}
      & \citet{paulfrank:2019}\textsuperscript{*} & 59.66 & 51.33 & 55.18\\

      \cline{2-5}
      & BERT  & 65.63 & 56.91 & 60.96 \\ 

      \hline
      \hline
      & \multicolumn{4}{l|} {\textbf{Label Embeddings}}\\
      \cline{2-5}
       
      & LEAM w/ GloVe & 59.81 & 54.46 & 57.03 \\

      & LEAM w/ BERT Features & 67.29 & 54.48 & 60.22 \\
      \cline{2-5}
      Adding & BERT + Labels as Input  & 63.05 & 61.70 & 62.36 \\ \cline{2-5}

    Label  Semantics & \multicolumn{4}{l|} {\textbf{Label Correlation}}\\
    \cline{2-5}

      & Learned Correlations & 56.50 & 71.47 & 63.11\\
      \cline{2-5}
      & Semi-supervision & 57.94 & 76.35 & 65.88\\
    \hline
    \end{tabular}
\caption{Comparison Results on ROCStories with Plutchik emotion labels}\label{tab:mainresults}

\end{table*}

\section{Experimental Setup}
\label{sec:baseline}

\noindent We compare our proposed models with the models presented in \citet{rashkin2018modeling}, the LEAM architecture of \citet{wang2018joint}, and fine-tuned BERT models \cite{devlin-etal-2019-bert} for multi-label classification without label semantics. For all the models we report the micro-averaged Precision, Recall and F1 score of the emotion prediction task.

\citet{rashkin2018modeling} modeled character context and pre-trained on free response data to predict the mental states of characters using different encoder-decoder setups, including BiLSTMs, CNNs, the recurrent entity network (REN) \cite{henaff2016tracking}, and neural process networks (NPN) \cite{bosselut2017simulating}.
Additionally, we compare with the self-attention architecture proposed in \cite{paulfrank:2019}, without the knowledge from ConceptNet \cite{speer-havasi-2012-representing} and ELMo embeddings \cite{Peters2018DeepCW}.

To compare against LEAM, we compare it against our proposal of the LEAM+BERT model, where our label attention is computed from BERT representations of each of the label sentences, and words in the input sentence. We also encode the sentence and context separately in a BiLSTM layer as done in \citet{rashkin2018modeling}.

We also fine-tuned a BERT-base-uncased model for emotion classification, using $x_{s}$, $x_{c}$ and $L_{s}$ as inputs. This beats the other baselines by a significant margin, and is thus a strong new baseline. All our models are evaluated on the emotion reaction prediction task over the eight emotion labels (Plutchik categories) annotated in the \citet{rashkin2018modeling} dataset. We follow their evaluation setup, and report the final results on the test set. We use pre-trained GloVe embeddings (100d) and BERT-base-uncased representations with the LEAM model. The final classifier used in all models is a feed-forward layer, followed by a sigmoid.

%% file: acl2020-templates/results.tex
\section{Results}
Table~\ref{tab:mainresults} compares the performance of the baselines with our models that use label semantics. Among the baselines, the fine-tuned BERT base model obtains the best results. Adding label embeddings (section 3.1) to the basic BiLSTM via LEAM model provides substantial increase, more than 27 absolute points in F1. We swapped in BERT features instead of GloVe and found a further 3 point improvement. 
The BERT baseline beat both of these, but appending label sentences as additional input to fine-tuned BERT increased its performance by 1.4 F1 points.

\begin{table*}[ht!]
    
    \begin{tabular}{c|c|c|c}\hline
    \toprule
      \textbf{Sentence} & \textbf{Ground Truth} & \textbf{LS} & \textbf{NoLS} \\
        
      \hline
      \toprule
      \multirow{2}{*}{And nobody could give him any direction} & {Sad, Disgust} & {Sad, Disgust } &  \multirow{2}{*}{Sad} \\
      & {Surprise} & {Anger} & \\ \hline
      \multirow{2}{*}{She said Mark can come for free} & {Joy, Trust} & {Joy ,Trust} & {Joy, Anticipation} \\
      &Anticipation&Anticipation &\\
      \hline
      \midrule
      \multirow{2}{*}{He is relieved that it was not harmed} & {Joy, Surprise } & {Joy, Surprise } & {Fear, Surprise } \\
      &Anticipation&Anticipation&\\\hline
      
      The marshmallows were totally smooshed & {Anger, Sad} & {Anger, Sad} & {Joy, Anticipation} \\

      \hline

      \bottomrule
      
    \end{tabular}

 \caption{Prediction of labels with label semantics (LS) versus without label semantics (NoLS). Including label semantics helps the model predict semantically labels (high correlations), with high probability. }
 \label{tab:model_predict}

\end{table*}
A further increase of 2 points in F1 is achieved by tracking label-label correlations through training loss and inference logits. In addition, adding semi-supervision yields the best gain of more than 4.9 points in F1 over basic BERT, providing a significant advance in state-of-the-art results for emotion inference in this dataset. We also checked the statistical significance of the Semi-supervision model (Table \ref{tab:mainresults}) against the Learned Correlations, BERT+Labels as Input, LEAM w/ BERT Features and the BERT model using the Randomization Test \cite{smucker2007comparison}. This involved comparing the outputs of the Semi-supervision model with the above mentioned models after creating 100,000 random permutations. The Semi-supervision model achieved statistically significant improvement over all the baselines. 

We did further qualitative analysis of the results on the dev set to better understand the performance of the Semi-supervised Label Semantics model. Compared to base BERT, this model predicts more emotion classes per instance (8839 vs 5024). \eat{The distinction between the correct and incorrect labels are much higher than the simple BERT model, which highlights the higher confidence in predictions. and more correct emotion classes per instance (4421 vs 3665), and has fewer semantically opposite predictions to ground truth \nb{( vs )}}
The wrong predictions of this model have lower probabilities than the correct labels suggesting that classification could be further improved with proper threshold identification. This model is also better at capturing the semantic relations between labels during prediction. This is highlighted through some examples in Table \ref{tab:model_predict}.

%% file: acl2020-templates/related.tex
\section{Related Work}
One of the most widely-used work in narrative understanding introduced ROCStories, a dataset for evaluating story understanding \cite{mostafazadeh2016corpus}. On a subset of these stories \cite{rashkin2018modeling} added annotations for causal links between events in stories and mental states of characters. They model entity state to predict emotional reactions and motivations for causing events occurring in ROCStories. Additionally, they also introduce a new dataset annotation that tracks emotional reactions and motivations of characters in stories. Other work looked at encoding external knowledge sources to augment motivation inference \cite{paulfrank:2019} on the same dataset. Both treat labels as anonymous classes, whereas this work explores modeling the semantics of the emotion labels explicitly.
Recent work in multi-label emotion classification has shown that using the relation information between labels can improve performance.
\cite{kurata2016improved} use the label co-occurrence information in the final layer of the neural network to improve multi-label classification. Correlation-based label representations have also been used for music classification styles ~\cite{zhao-etal-2019-review}. Our work builds on these and adds a similar result showing that label correlations can have significant impact for emotion label inference.

%% file: acl2020-templates/conclusions.tex
\section{Conclusions}
We present new results for the multi-label emotion classification task of \citet{rashkin2018modeling}, extending previous reported results by 10.7 F1 points (55.1 to 65.8). 
The multi-label nature of emotion prediction  lends itself naturally to use the correlations between the labels themselves. 
Further, we showed that modeling the class labels as semantic embeddings helped to learn better representations with more meaningful predictions.
As with many tasks, BERT provided additional context, but our integration of these label semantics showed significant improvements.
We believe these models can improve many other NLP tasks where the class labels carry inherent semantic meaning in their names.